\documentclass[runningheads]{llncs}
\usepackage[T1]{fontenc}
\usepackage{graphicx,verbatim}
\usepackage{amsmath}
\usepackage{multicol,multirow}
\begin{document}
\title{Distribution-Based Masked Medical Vision-Language Model Using Structured Reports}
%
%\begin{comment}  %% Removed for anonymized MICCAI 2025 submission
\author{Shreyank N Gowda\inst{1} \and
Ruichi Zhang\inst{2} \and
Xiao Gu\inst{3} \and
Ying Weng\inst{4} \and
Lu Yang\inst{2}}
\authorrunning{S. N. Gowda et al.}
\institute{School of Computer Science, University of Nottingham, NG8 1BB
Nottingham, U.K. \and
Department of Computer Science and Technology, School of Informatics, Xiamen University, Xiamen, 361005, China \and CHI Lab, University
of Oxford, OX3 7DQ Oxford, U.K \and School of Computer Science, University of Nottingham Ningbo China, Ningbo, 315100, China \\
\email{shreyank.narayanagowda@nottingham.ac.uk}}

%\end{comment}

\maketitle              % typeset the header of the contribution
\begin{abstract}
Medical image-language pre-training aims to align medical images with clinically relevant text to improve model performance on various downstream tasks. However, existing models often struggle with the variability and ambiguity inherent in medical data, limiting their ability to capture nuanced clinical information and uncertainty. This work introduces an uncertainty-aware medical image-text pre-training model that enhances generalization capabilities in medical image analysis. Building on previous methods and focusing on Chest X-Rays, our approach utilizes structured text reports generated by a large language model (LLM) to augment image data with clinically relevant context. These reports begin with a definition of the disease, followed by the `appearance' section to highlight critical regions of interest, and finally `observations' and `verdicts' that ground model predictions in clinical semantics. By modeling both inter- and intra-modal uncertainty, our framework captures the inherent ambiguity in medical images and text, yielding improved representations and performance on downstream tasks. Our model demonstrates significant advances in medical image-text pre-training, obtaining state-of-the-art performance on multiple downstream tasks.

\keywords{Vision-Language  \and Uncertainty \and Chest X-Ray}
% Authors must provide keywords and are not allowed to remove this Keyword section.

\end{abstract}

\section{Introduction}
With rapid advancements in deep learning, computer-aided diagnosis in medicine has seen significant progress across various model architectures. However, these models are often trained on specific anatomical or disease categories, requiring expensive data annotation and re-training when applied to new diseases, which limits their broader applicability. Although deep learning has thrived on large-scale labeled datasets from natural images~\cite{resnet,gowda2018colornet}, annotating medical images is a much more time-intensive and costly process. A typical approach involves pre-training on extensive datasets like ImageNet~\cite{imagenet} and fine-tuning on specialized medical datasets~\cite{wen2021rethinking}. However, this method often struggles to achieve generalized performance due to the significant domain gap.

Medical image analysis stands as a critical area in healthcare, where accurate interpretation can significantly impact clinical outcomes. Traditional methods in medical imaging rely heavily on annotated datasets~\cite{wen2021rethinking}, which are costly and time-consuming to curate, especially for new or rare diseases. Recent advances in self-supervised pre-training methods like contrastive predictive coding~\cite{oord2018representation} and masked language modeling~\cite{devlin2018bert} have shown promise in leveraging large, unlabeled datasets to learn robust image and text representations. While general vision-language models like CLIP~\cite{radford2021learning} have achieved impressive performance on natural images, they struggle with medical data due to domain-specific language and visual features~\cite{medklip,gowda2024masks}. Existing medical image-text approaches like ConVIRT~\cite{convirt}, PRIOR~\cite{prior}, M\& M~\cite{gowda2024masks} and GLoRIA~\cite{gloria} often overlook the inherent uncertainties present in medical data, where variability in clinical descriptions and visual cues can lead to ambiguous interpretations. Whilst uncertainty has been explored in a wide variety of contexts~\cite{ghesu2021quantifying,yang2021uncertainty,gowda2024cc,baumgartner2019phiseg}, to the best of our knowledge we are the first to explore this on chest X-ray image-text pre-training.

We propose an uncertainty-aware pre-training model for medical image-text data, focusing on X-ray data. We leverage Distribution-based Masked Image-Language Modeling (D-MLM) to capture both inter- and intra-modal uncertainties, thus enabling more nuanced understanding and alignment between images and associated text. By treating representations as probabilistic distributions rather than deterministic points, D-MLM allows the model to capture the natural ambiguity and variability in medical data, enhancing its capacity for accurate and robust prediction.
Since existing reports have semantic inconsistencies~\cite{medklip,gowda2024masks}, a key component of our approach involves the structured text reports generated by a large language model (LLM)~\cite{gpt4}. We first follow M\&M~\cite{gowda2024masks} that takes the original reports and converts them to a series of `Observations' and `Verdicts'. To this, we add at the beginning a definition of the disease, followed by an `Appearance' section to guide attention to critical regions in the image, and ending with `Observations' and `Verdicts' that offer conclusive insights. This structured report provides clinically relevant context that anchors the model's predictions, ensuring that outputs align with medical semantics. We show that using such a structured report significantly improves our overall performance. We use these reports along with their corresponding images to do the pre-training. We show using our approach improves performance on multiple different downstream tasks and different benchmarks.
The contributions of this work are threefold: 
\begin{itemize}
\item We introduce D-MLM to effectively model multimodal uncertainty in medical image-text data, enhancing the robustness of pre-trained representations. 
\item We leverage structured reports to provide clinically meaningful context for model predictions, aligning with medical semantics. 
\item Through extensive experiments on downstream tasks, we demonstrate the superiority of our method over traditional deterministic approaches, setting a new standard for multimodal pre-training in the chest X-ray domain.
\end{itemize}

\section{Method}

\begin{figure*}[ht]
    \centering
    \includegraphics[width=\textwidth]{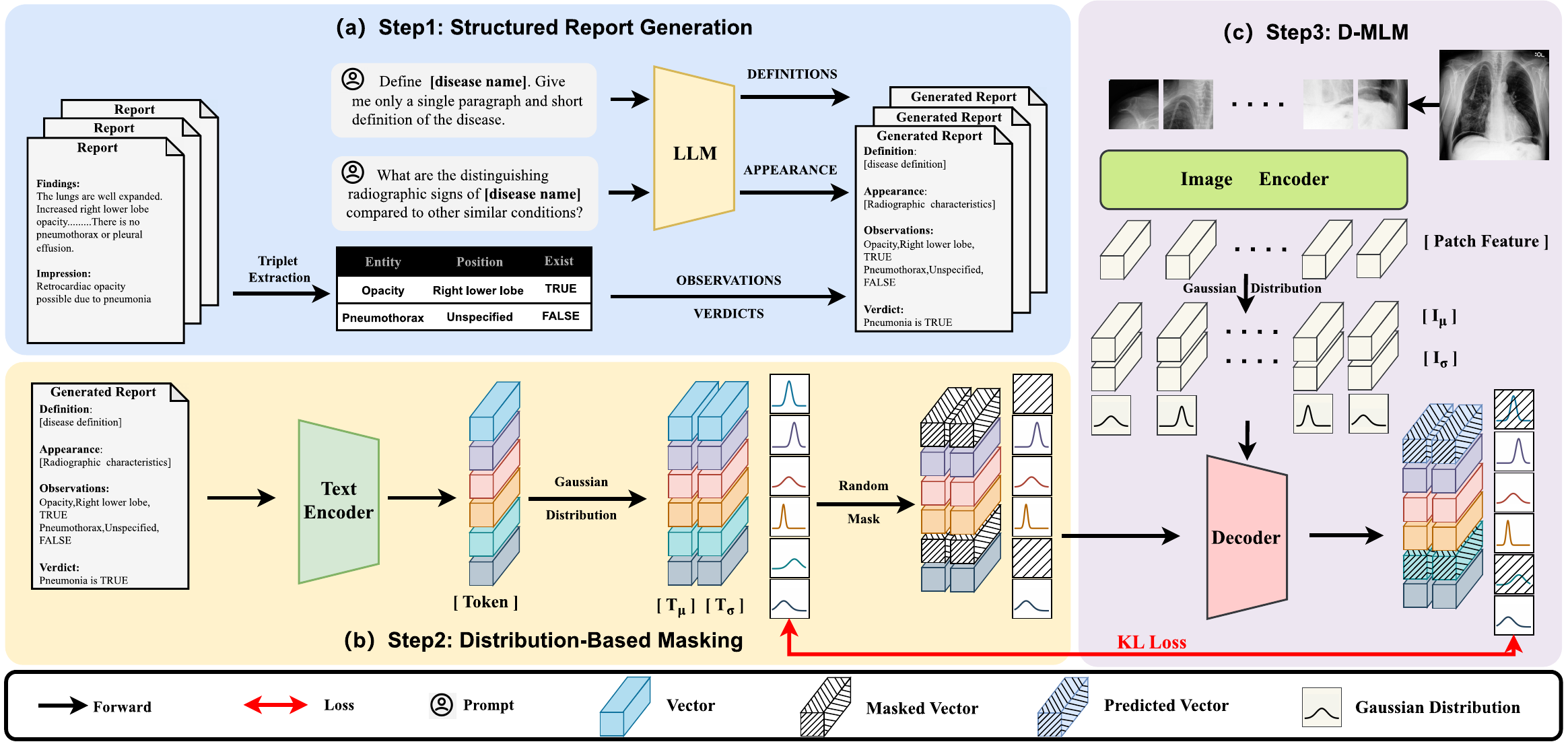}
    \caption{
        Overview of our D-MLM framework for medical image-text alignment. (a) Structured Report Generation: An LLM creates standardized reports with disease definitions, appearance guidance, observations, and verdicts. (b) Distribution-Based Representation: Image and text features are encoded as probabilistic distributions with means and variances. (c) D-MLM: The model masks and predicts tokens and patches using distribution-based techniques, optimizing with KL divergence loss to enhance uncertainty modeling.
    }
    \label{fig:D-MLM_overview}
\end{figure*}

This section presents our uncertainty-aware pre-training framework for medical image-text alignment. Our Distribution-based Masked Image-Language Modeling (D-MLM) approach combines LLM-generated structured reports for clinical context with probabilistic representations that capture inherent medical data ambiguity. Figure~\ref{fig:D-MLM_overview} provides an overview.

\subsection{Structured Report Generation}

We generate structured reports for each medical image using an LLM, following a standardized format with three key components.

\begin{itemize}
    \item \textbf{Definition:} This section provides a concise description of the disease or clinical condition under consideration. To generate this content, we prompt the LLM with, \textit{
``Define [disease name]. Give me only a single paragraph and short definition of the disease.''
} and the model returns a standardized response as, \textit{``Definition: [disease definition].''} This provides a concise introduction to the condition, grounding the model in clinically accurate language.

    \item \textbf{Appearance:} This section describes key radiographic features and diagnostic areas, by prompting the LLM with, \textit{``What are the distinguishing radiographic signs of [disease name] compared to other similar conditions?''} The model then responds with \textit{``Radiographic characteristics: [disease-specific radiographic characteristics]''}, guiding the model’s focus to relevant visual cues within the image data.

    \item \textbf{Observations and Verdicts:} This part details visual findings and clinical verdicts, anchoring predictions in medical context. Following Masks \& Manuscripts~\cite{gowda2024masks}, it emphasizes structured clinical reasoning for text-image alignment.
\end{itemize}

By following this format, the structured reports enhance consistency in text inputs, reduce variability in clinical descriptions, and support the model in achieving precise alignment between image and text features. Details of the definitions and appearances can be found in the link: \newline \url{https://github.com/kini5gowda/MIMIC-CXR-text}

\subsection{Distribution-Based Masked Image-Language Modeling (D-MLM)}

Our approach centers on Distribution-based Masked Image-Language Modeling (D-MLM), which represents image and text as probabilistic distributions to capture inter-modal and intra-modal uncertainty in medical data.

In D-MLM, image $\mathbf{I}$ is encoded through ImageNet-pretrained ViT-B~\cite{vit}, while text $\mathbf{T}$ uses ClinicalBERT~\cite{cbert}. Both outputs are transformed into multivariate Gaussian distributions, with each token or patch represented as:
\begin{equation}
   h_{i} = N(\mu_{i}, \sigma_{i}^2)
\end{equation}
where $\mu_{i}$ and $\sigma_{i}^2$ are mean and variance vectors. This distribution-based approach captures data variability better than fixed-point representations.

For training, we mask 30\% of text tokens (higher than the standard 15\%~\cite{devlin2018bert}) and focus image masking on diagnostically relevant regions identified from the 'Appearance' section of reports. This adaptive masking strategy emphasizes clinically significant features.

\subsection{Pre-Training Objective: Distribution-Based Masked Image-Language Modeling}

The pre-training objective for D-MLM optimizes the model's ability to reconstruct masked elements using both modalities, framed as a probabilistic reconstruction task. For a masked token or patch $h_{i}$, the model predicts:
\begin{equation}
   p(h_{i} | \mathbf{I}, \mathbf{T}_{\backslash i}) = N(\hat{\mu}_{i}, \hat{\sigma}_{i}^2)
\end{equation}

The loss function uses Kullback-Leibler (KL) divergence between predicted and ground truth distributions:
\begin{equation}
   \begin{split}
       \mathcal{L}_{\text{D-MLM}} = \mathbf{E}_{(\mathbf{I}, \mathbf{T}) \sim D} \Bigg[ 
       \sum_{t \in \mathcal{M}_{\text{text}}} \text{KL}\left( N(\hat{\mu}_{t}, \hat{\sigma}_{t}^2) \parallel N(\mu_{t}, \sigma_{t}^2) \right) \\
       + \sum_{p \in \mathcal{M}_{\text{image}}} \text{KL}\left( N(\hat{\mu}_{p}, \hat{\sigma}_{p}^2) \parallel N(\mu_{p}, \sigma_{p}^2) \right) \Bigg]
   \end{split}
\end{equation}
where $\mathcal{M}_{\text{text}}$ and $\mathcal{M}_{\text{image}}$ are the sets of masked tokens and patches.

\paragraph{Uncertainty-Aware Alignment Loss.} We introduce an alignment loss that minimizes Wasserstein distance between probabilistic embeddings of aligned image-text pairs:
\begin{equation}
   \begin{split}
       \mathcal{L}_{\text{align}} = \sum_{(h_{i}^{\{\text{text}\}}, h_{j}^{\{\text{image}\}}) \in \mathcal{A}} 
       W\Big(N(\mu_{i}^{\{\text{text}\}}, \sigma_{i}^{\{\text{text}\}^2}), \\
       N(\mu_{j}^{\{\text{image}\}}, \sigma_{j}^{\{\text{image}\}^2})\Big)
   \end{split}
\end{equation}
where $\mathcal{A}$ is the set of aligned pairs and $W(\cdot)$ denotes Wasserstein distance. Our ablation studies show this slightly improves performance, though even without it we outperform existing approaches.

\paragraph{Overall Loss Function.} The total pre-training loss combines:
\begin{equation}
   \mathcal{L}_{\text{total}} = \lambda\mathcal{L}_{\text{D-MLM}} + (1- \lambda) \mathcal{L}_{\text{align}}
\end{equation}
where $\lambda$ balances the contributions of both losses.

\section{Experimental Analysis}

\subsection{Datasets}
In this work, we use several publicly available chest X-ray datasets that have been commonly adopted in recent research~\cite{gloria,prior,medklip,gowda2024masks}. \textbf{MIMIC-CXR v2~\cite{johnson2019mimic}} includes 377,110 chest radiographs linked to 227,835 imaging studies, annotated with 14 common chest conditions, which we leverage for pre-training our model. \textbf{RSNA Pneumonia Detection~\cite{shih2019augmenting}} contains approximately 30,000 chest X-rays with bounding box annotations for pneumonia, which we split 60/20/20 for training, validation, and testing. \textbf{SIIM-ACR Pneumothorax~\cite{pneumo}} comprises 12,954 chest X-rays with image-level pneumothorax annotations and segmentation masks where present; we use a 60/20/20 split for classification tasks and focus on 2,669 samples with pneumothorax for segmentation. \textbf{NIH Chest X-Ray Dataset~\cite{nih}} contains 112,120 frontal-view X-rays from 30,805 patients annotated with 14 thoracic conditions, divided 80/10/10 for training, validation, and testing. \textbf{CheXpert~\cite{chexpert}} consists of 224,316 X-ray images from 65,240 patients, automatically labeled for 14 thoracic observations, supporting multi-label classification; we split the training data 80/20 and use the official validation set for testing. \textbf{COVIDx CXR~\cite{covid}} includes 29,986 X-rays from 16,648 patients labeled for COVID-19 diagnosis, which we split 70/20/10. \textbf{Edema Severity~\cite{edema}}, derived from MIMIC-CXR, comprises 6,524 X-ray images with pulmonary edema severity scores from 0 to 3, which we split 60/20/20 for fine-grained classification.

\subsection{Classification}

\subsubsection{Semi- and Fully-Supervised}

We conduct both semi-supervised and fully supervised classification experiments across three datasets: RSNA Pneumonia, SIIM-ACR, and CheXpert. The experiments vary the proportion of labeled data from 1\% to 100\%. For all methods, we report results based on averages and standard deviations over five runs, as provided by PRIOR~\cite{prior}. The results, presented in Table~\ref{tab:supervised_classification}, demonstrate that D-MLM surpasses previous methods by up to 2.3\%.

\begin{table*}
    \centering
\resizebox{0.88\textwidth}{!}{%
\begin{tabular}{cccccccccc}
\hline    & \multicolumn{3}{c}{ RSNA Pneumonia } & \multicolumn{3}{c}{ SIIM-ACR } & \multicolumn{3}{c}{ CheXpert } \\
 Methods & $1 \%$ & $10 \%$ & $100 \%$ & $1 \%$ & $10 \%$ & $100 \%$ & $1 \%$ & $10 \%$ & $100 \%$ \\
\hline MoCo~\cite{moco} & $82.33$ & $85.22$ & $87.90$ & $75.49$ & $81.01$ & $88.43$ & $78.00$ & $86.27$ & $87.24$ \\
SimCLR~\cite{chen2020simple} & $80.18$ & $84.60$ & $88.07$ & $74.97$ & $83.21$ & $88.72$ & $67.41$ & $86.74$ & $87.97$ \\
\hline
ConVIRT~\cite{convirt} & $83.98$ & $85.62$ & $87.61$ & $84.17$ & $85.66$ & $91.50$ & $85.02$ & $87.58$ & $88.21$ \\
GLoRIA~\cite{gloria} & $84.12$ & $86.83$ & $89.13$ & $85.05$ & $88.51$ & $92.11$ & $83.61$ & $87.40$ & $88.34$ \\
BioViL~\cite{biovil} & $81.95$ & $85.37$ & $88.62$ & $79.89$ & $81.62$ & $90.48$ & $80.77$ & $87.56$ & $88.41$ \\
LoVT~\cite{lovt} & $85.51$ & $86.53$ & $89.27$ & $85.47$ & $88.50$ & $92.16$ & $85.13$ & $88.05$ & $88.27$ \\
PRIOR~\cite{prior} & $85.74$ & $87.08$ & $89.22$ & $87.27$ & $89.13$ & $92.39$ & $86.16$ & $88.31$ & $88.61$ \\
MedKLIP~\cite{medklip} & $87.31$ & $87.99$ & $89.31$ & $85.27$ & $90.71$ & $91.88$ & $86.24$ & $88.14$ & $88.68$ \\
M\&M~\cite{gowda2024masks} & $88.11$ & $89.44$ & $91.91$ & $88.81$ & $91.15$ & $93.88$ & $88.45$ & $90.02$ & $90.88$ \\
MLIP~\cite{mlip} & 89.30 & 90.04 & 90.81 & - & - & - & 89.03 & 89.44 & 90.04 \\
UniMedI~\cite{unimedi} & 90.02 & 90.41 & 91.47 & - & - & - & 89.44 & 89.72 & 90.51 \\
IMITATE~\cite{imitate} & 91.73 & 92.85 & 93.46 & - & - & - & 89.13 & 89.49 & 89.66 \\
\textbf{D-MLM (Ours)} & \textbf{91.94} & \textbf{92.91} & \textbf{93.84} & \textbf{91.11} & \textbf{92.44} & \textbf{95.18} & \textbf{89.80} & \textbf{90.41} & \textbf{91.45} \\
\hline
\end{tabular}}
\caption{Comparison of semi-supervised and supervised classification results after fine-tuning on RSNA~\cite{shih2019augmenting}, SIIM~\cite{pneumo}, and CheXpert~\cite{chexpert}. Methods are trained on 1\%-100\% of training data and evaluated using AUC-ROC.}
    \label{tab:supervised_classification}
\end{table*}

\vspace{-10mm}

\begin{table*}[ht]
\centering
\begin{tabular}{lccccccccc}
\hline
& \multicolumn{3}{c}{RSNA Pneumonia} & \multicolumn{3}{c}{SIIM-ACR} & \multicolumn{3}{c}{NIH Chest X-Ray} \\
Methods & AUC↑ & F1↑ & ACC↑ & AUC↑ & F1↑ & ACC↑ & AUC↑ & F1↑ & ACC↑ \\
\hline
ConVIRT~\cite{convirt} & 80.42 & 58.42 & 76.11 & 64.31 & 43.29 & 57.00 & 61.01 & 16.28 & 71.02 \\
GLoRIA~\cite{gloria} & 71.45 & 49.01 & 71.29 & 53.42 & 38.23 & 40.47 & 66.10 & 17.32 & 77.00 \\
BioViL~\cite{biovil} & 82.80 & 58.33 & 76.69 & 70.79 & 48.55 & 69.09 & 69.12 & 19.31 & 79.16 \\
PRIOR~\cite{prior} & 85.58 & 62.91 & 77.85 & 86.62 & 70.11 & 84.44 & 74.51 & 23.29 & 84.41 \\
MedKLIP~\cite{medklip} & 86.94 & 63.42 & 80.02 & 89.24 & 68.33 & 84.28 & 76.76 & 25.25 & 86.19 \\
M\&M~\cite{gowda2024masks} & 88.91 & 66.58 & 83.14 & 91.15 & 71.58 & 86.15 & 77.92 & 27.55 & 88.52 \\
\textbf{D-MLM (Ours)} & \textbf{90.15} & \textbf{68.42} & \textbf{85.11} & \textbf{91.45} & \textbf{72.18} & \textbf{86.88} & \textbf{79.54} & \textbf{28.81} & \textbf{90.15} \\
\hline
\end{tabular}
\caption{ Comparing recent state-of-the-art methods on zero-shot classification task. We use AUC, F1 and ACC scores for comparison. Following MedKLIP~\cite{medklip} for evaluation on NIH Chest X-Ray, the metrics all refer to the macro average on the 14 diseases.}
\label{table:zero_class}
\end{table*}

\vspace{-10mm}

\subsubsection{Zero-Shot}

We assess zero-shot classification performance of state-of-the-art models on RSNA Pneumonia, SIIM-ACR, and NIH Chest X-Ray datasets, evaluating generalization to 'seen' conditions from different clinical sources. Following MedKLIP~\cite{medklip}, we categorize this as zero-shot classification rather than domain adaptation. Table~\ref{table:zero_class} shows our approach outperforming prior methods by up to 1.96\% across metrics when evaluated directly after MIMIC-CXR pre-training.

Additionally, we test the model on an entirely new disease, COVID-19, which is absent in the pre-training data. As shown in Table~\ref{tab:covid_description_comparison}, our approach achieves improvements of up to 2.04\%.

\subsection{Grading}

Beyond diagnosis, assessing disease severity is essential. We fine-tune our pre-trained features on the Edema Severity~\cite{edema} dataset, which classifies conditions on a 0-3 scale. Table~\ref{tab:grade} shows average scores across all severity levels.

\begin{table}[h!]
\begin{minipage}{0.48\textwidth}
\centering
\begin{tabular}{cccc}
\hline
Methods & AUC↑ & F1↑ & ACC↑ \\
\hline
ConVIRT~\cite{convirt} & 52.08 & 69.02 & 52.66 \\
GloRIA~\cite{gloria}  & 66.59 & 70.07 & 60.83 \\
BioViL~\cite{biovil}   & 53.82 & 69.10 & 53.75 \\
MedKLIP~\cite{medklip} & 73.96 & 76.70 & 70.06 \\
M\&M~\cite{gowda2024masks} & 75.15 & 77.89 & 73.35 \\
\textbf{D-MLM (Ours)} & \textbf{77.19} & \textbf{79.52} & \textbf{74.78} \\
\hline
\end{tabular}
\caption{Performance comparison for Zero-Shot Classification on Covid-19 CXR. We use AUC, F1 and ACC scores for comparison.}
\label{tab:covid_description_comparison}
\end{minipage}
\hfill
\begin{minipage}{0.48\textwidth}
\centering
\begin{tabular}{lccc}
\hline
Methods & AUC↑ & F1↑ & ACC↑ \\
\hline
ConVIRT~\cite{convirt}   & 77.00 & 56.76 & 69.19   \\
GLoRIA~\cite{gloria}    & 77.74 & 57.98 & 71.45  \\
BioViL~\cite{biovil}     & 75.40 & 55.72 & 69.14   \\
MedKLIP~\cite{medklip}     & 78.98 & 58.26 & 72.80   \\
M\&M~\cite{gowda2024masks} & 80.71 & 60.18 & 73.91\\
\textbf{D-MLM (Ours)} & \textbf{82.93} & \textbf{62.11} & \textbf{75.51} \\
\hline
\end{tabular}
\caption{Comparison with state-of-the-art methods on fine-tuning edema severity grading multi-class classification task. Only the average across all classes has been reported here.}
\label{tab:grade}
\end{minipage}
\end{table}

\vspace{-10mm}

\subsection{Segmentation}

%\vspace{-5mm}

\begin{table*}
\centering

\begin{tabular}{lccccccccc}
\hline
\multirow{2}{*}{Methods} & \multicolumn{3}{c}{RSNA Pneumonia} & \multicolumn{3}{c}{SIIM-ACR} & \multicolumn{3}{c}{Covid-19} \\ \cline{2-10} 
                                   & 1\%      & 10\%     & 100\%    & 1\%        & 10\%       & 100\%      & 1\%      & 10\%     & 100\%    \\ \hline
Scratch                            & 43.47   & 60.47   & 70.68   & 21.33     & 33.23     & 74.47     & 14.81   & 23.67   & 32.28   \\
ConVIRT~\cite{convirt}                           & 57.06   & 64.91   & 72.01   & 54.06     & 61.21     & 73.52     & 19.95   & 27.24   & 37.37   \\
GLoRIA~\cite{gloria}                            & 65.55   & 69.07   & 73.28   & 56.73     & 57.78     & 76.94     & 18.89   & 28.09   & 38.69   \\
BioVil~\cite{biovil}                             & 68.24   & 70.38   & 72.49   & 62.67     & 69.98     & 78.49     & 21.13   & 32.39   & 41.62   \\
PRIOR~\cite{prior}                               & 70.11   & 70.88   & 74.43   & 66.14     & 71.24     & 78.85   & 23.66  & 34.72   & 43.01   \\ 
MedKLIP~\cite{medklip}                               & 70.64   & 71.62   & 75.79   & 66.59     & 72.10     & 79.37    & 24.45   & 35.39   & 43.99   \\
M\&M~\cite{gowda2024masks} & 72.28 & 73.11 & 76.68 & 69.55 & 73.47 & 80.28 & 28.25 & 37.32 & 45.04 \\
UniMedI~\cite{unimedi} & 67.80 & 73.10 & 75.30 & - & - & - & - & - & - \\
MLIP~\cite{mlip} & 67.70 & 68.80 & 73.50 & 51.60 & 60.80 & 68.10 & - & - & - \\
\textbf{D-MLM (Ours)} & \textbf{74.11} & \textbf{74.77} & \textbf{77.16} & \textbf{70.95} & \textbf{74.49} & \textbf{81.12} & \textbf{30.41} & \textbf{38.11} & \textbf{46.11} \\
\hline
\end{tabular}
\caption{Comparing Dice scores with other state-of-the-art methods on segmentation tasks. We report on three diseases with varying percentages of labeled data 1\%, 10\%, 100\% and see improvements in all cases.}
\label{tab:seg}
\end{table*}

%\vspace{-10mm}

Table~\ref{tab:seg} presents our fine-tuning experiments for segmenting three distinct diseases, where we utilize 1\%, 10\%, and 100\% of the available data. Regardless of the varying image distributions associated with each disease, our techniques consistently outperform current leading methods. We see significant gains in particular when data is scarce outperforming previous works by up to 2.16\%.

\subsection{Implementation Details}

To ensure fair comparison, we use a ViT-B~\cite{vit} image backbone pre-trained on ImageNet~\cite{imagenet}, with images resized to 224$\times$224, and ClinicalBERT~\cite{cbert} as the text backbone, both with a latent dimension of 768. Training is conducted with a batch size of 128 on 4 NVIDIA Tesla V100 GPUs, using AdamW with a weight decay of 0.05. Definitions and radiographic descriptions are generated by GPT-4 based on specific prompts. In our D-MLM framework, masking ratios are dynamically adjusted: an adaptive ratio for images based on the paired text, and a 30\% ratio for text to leverage image context. Pre-training is performed for 100 epochs, while fine-tuning occurs over 10 epochs. The learning rate is warmed up to 3$\times$10$^{-4}$ with a cosine scheduler, with encoder rates set to 10$^{-5}$.

\subsection{Ablation Study}

We evaluate key components of our approach through focused experiments on the NIH Chest X-Ray dataset in zero-shot settings.

\begin{table}[h!]
\begin{minipage}{0.48\textwidth}
\centering
\begin{tabular}{cccc}
\hline
\textbf{Methods} & \textbf{AUC↑} & \textbf{F1↑} & \textbf{ACC↑} \\
\hline
Original Report & 69.95 & 20.04 & 77.71 \\
Triplet & 73.48 & 24.42 & 82.89 \\
KE-Triplet & 76.84 & 26.11 & 86.55 \\
M\&M~\cite{gowda2024masks} & 77.92 & 27.55 & 88.52 \\
\textbf{Ours} & \textbf{79.54} & \textbf{28.81} & \textbf{90.15} \\
\hline
\end{tabular}
\caption{Ablation on reports.}
\label{tab:ablation_report}
\end{minipage}
\hfill
\begin{minipage}{0.48\textwidth}
\centering
\begin{tabular}{cccc}
\hline
\textbf{Methods} & \textbf{AUC↑} & \textbf{F1↑} & \textbf{ACC↑} \\
\hline
No Masking & 61.48 & 16.33 & 70.54 \\
MAE~\cite{masked}  & 68.84 & 18.85 & 75.59 \\
MaskVLM~\cite{maskvlm} & 58.87 & 14.96 & 66.69 \\
M\&M~\cite{gowda2024masks}  & 77.92 & 27.55 & 88.52 \\
\textbf{D-MLM} & \textbf{79.54} & \textbf{28.81} & \textbf{90.15} \\
\hline
\end{tabular}
\caption{Ablation on masking.}
\label{tab:ablation_dmlm}
\end{minipage}
\end{table}

\vspace{-10mm}

Our structured reports with definition and appearance sections improve performance over M\&M~\cite{gowda2024masks} and other baselines (Table~\ref{tab:ablation_report}). These sections provide richer clinical context, enhancing image-text alignment. Our D-MLM approach outperforms alternative masking strategies (Table~\ref{tab:ablation_dmlm}), demonstrating the value of modeling features as probabilistic distributions. We use a fixed masking ratio of 0.3 and a $\lambda$ of 0.2 based on experimental analysis. Notably, even without alignment loss, our model outperforms competitors, indicating that while beneficial, alignment loss is not essential for state-of-the-art performance.

\iffalse
\begin{table}[!t]
\centering
\begin{tabular}{cccc}
\hline
\textbf{$\lambda$} & \textbf{AUC↑} & \textbf{F1↑} & \textbf{ACC↑} \\
\hline
0 & 79.41 & 28.45 & 89.95 \\
0.1 & 79.47 & 28.78 & 90.11 \\
0.2 & \textbf{79.54} & \textbf{28.81} & \textbf{90.15} \\
0.4 & 79.10 & 27.51 & 88.55 \\
0.8 & 73.61 & 23.23 & 81.51 \\
1 & 68.65 & 19.22 & 73.41 \\
\hline
\end{tabular}
\caption{Ablation study on choice of manuscript generation. We only report on the zero-shot setting as this shows us generalization ability. The dataset used is NIH Chest X-Ray. }
\label{tab:lambda}
\end{table}
\fi

\section{Conclusion}

In this work, we introduced Distribution-based Masked Image-Language Modeling (D-MLM), a novel approach for uncertainty-aware alignment of medical image-text data. By representing both image and text features as probabilistic distributions, D-MLM effectively captures the inherent ambiguity and variability in clinical data, allowing for robust and interpretable multimodal representations. Our method leverages structured reports, dynamically guided masking, and an uncertainty-aware alignment loss to enhance the model’s ability to learn meaningful associations between visual and textual information. Extensive experiments demonstrate that D-MLM achieves state-of-the-art performance across multiple medical tasks, highlighting its potential as a foundation for various downstream applications in healthcare. %Future work can explore extending D-MLM to other medical modalities and incorporating additional domain-specific knowledge to further improve performance and clinical relevance.

\bibliographystyle{splncs04}
\bibliography{main}

\begin{thebibliography}{10}
\providecommand{\url}[1]{\texttt{#1}}
\providecommand{\urlprefix}{URL }
\providecommand{\doi}[1]{https://doi.org/#1}

\bibitem{gpt4}
Achiam, J., Adler, S., Agarwal, S., Ahmad, L., Akkaya, I., Aleman, F.L., Almeida, D., Altenschmidt, J., Altman, S., Anadkat, S., et~al.: Gpt-4 technical report. arXiv preprint arXiv:2303.08774  (2023)

\bibitem{cbert}
Alsentzer, E., Murphy, J.R., Boag, W., Weng, W.H., Jin, D., Naumann, T., McDermott, M.: Publicly available clinical bert embeddings. arXiv preprint arXiv:1904.03323  (2019)

\bibitem{baumgartner2019phiseg}
Baumgartner, C.F., Tezcan, K.C., Chaitanya, K., H{\"o}tker, A.M., Muehlematter, U.J., Schawkat, K., Becker, A.S., Donati, O., Konukoglu, E.: Phiseg: Capturing uncertainty in medical image segmentation. In: International Conference on Medical Image Computing and Computer-Assisted Intervention. pp. 119--127. Springer (2019)

\bibitem{biovil}
Boecking, B., Usuyama, N., Bannur, S., Castro, D.C., Schwaighofer, A., Hyland, S., Wetscherek, M., Naumann, T., Nori, A., Alvarez-Valle, J., et~al.: Making the most of text semantics to improve biomedical vision--language processing. In: European Conference on Computer Vision. pp. 1--21. Springer (2022)

\bibitem{edema}
Chauhan, G., Liao, R., Wells, W., Andreas, J., Wang, X., Berkowitz, S., Horng, S., Szolovits, P., Golland, P.: Joint modeling of chest radiographs and radiology reports for pulmonary edema assessment. In: Medical Image Computing and Computer Assisted Intervention--MICCAI 2020: 23rd International Conference, Lima, Peru, October 4--8, 2020, Proceedings, Part II 23. pp. 529--539. Springer (2020)

\bibitem{chen2020simple}
Chen, T., Kornblith, S., Norouzi, M., Hinton, G.: A simple framework for contrastive learning of visual representations. In: International Conference on Machine Learning. pp. 1597--1607. PMLR (2020)

\bibitem{prior}
Cheng, P., Lin, L., Lyu, J., Huang, Y., Luo, W., Tang, X.: Prior: Prototype representation joint learning from medical images and reports. In: Proceedings of the IEEE/CVF International Conference on Computer Vision. pp. 21361--21371 (2023)

\bibitem{imagenet}
Deng, J., Dong, W., Socher, R., Li, L.J., Li, K., Fei-Fei, L.: Imagenet: A large-scale hierarchical image database. In: IEEE Conference on Computer Vision and Pattern Recognition. pp. 248--255. IEEE (2009)

\bibitem{devlin2018bert}
Devlin, J., Chang, M.W., Lee, K., Toutanova, K.: Bert: Pre-training of deep bidirectional transformers for language understanding. arXiv preprint arXiv:1810.04805  (2018)

\bibitem{vit}
Dosovitskiy, A., Beyer, L., Kolesnikov, A., Weissenborn, D., Zhai, X., Unterthiner, T., Dehghani, M., Minderer, M., Heigold, G., Gelly, S., et~al.: An image is worth 16x16 words: Transformers for image recognition at scale. arXiv preprint arXiv:2010.11929  (2020)

\bibitem{ghesu2021quantifying}
Ghesu, F.C., Georgescu, B., Mansoor, A., Yoo, Y., Gibson, E., Vishwanath, R., Balachandran, A., Balter, J.M., Cao, Y., Singh, R., et~al.: Quantifying and leveraging predictive uncertainty for medical image assessment. Medical Image Analysis  \textbf{68},  101855 (2021)

\bibitem{gowda2024cc}
Gowda, S.N., Clifton, D.A.: Cc-sam: Sam with cross-feature attention and context for ultrasound image segmentation. In: European Conference on Computer Vision. pp. 108--124. Springer (2024)

\bibitem{gowda2024masks}
Gowda, S.N., Clifton, D.A.: Masks and manuscripts: Advancing medical pre-training with end-to-end masking and narrative structuring. In: International Conference on Medical Image Computing and Computer-Assisted Intervention. pp. 426--436. Springer (2024)

\bibitem{gowda2018colornet}
Gowda, S.N., Yuan, C.: Colornet: Investigating the importance of color spaces for image classification. In: Asian conference on computer vision. pp. 581--596. Springer (2018)

\bibitem{masked}
He, K., Chen, X., Xie, S., Li, Y., Doll{\'a}r, P., Girshick, R.: Masked autoencoders are scalable vision learners. In: Proceedings of the IEEE/CVF Conference on Computer Vision and Pattern Recognition. pp. 16000--16009 (2022)

\bibitem{moco}
He, K., Fan, H., Wu, Y., Xie, S., Girshick, R.: Momentum contrast for unsupervised visual representation learning. In: Proceedings of the IEEE/CVF Conference on Computer Vision and Pattern Recognition. pp. 9729--9738 (2020)

\bibitem{resnet}
He, K., Zhang, X., Ren, S., Sun, J.: Deep residual learning for image recognition. In: Proceedings of the IEEE Conference on Computer Vision and Pattern Recognition (June 2016)

\bibitem{unimedi}
He, X., Yang, Y., Jiang, X., Luo, X., Hu, H., Zhao, S., Li, D., Yang, Y., Qiu, L.: Unified medical image pre-training in language-guided common semantic space. In: European Conference on Computer Vision. pp. 123--139. Springer (2025)

\bibitem{gloria}
Huang, S.C., Shen, L., Lungren, M.P., Yeung, S.: Gloria: A multimodal global-local representation learning framework for label-efficient medical image recognition. In: Proceedings of the IEEE/CVF International Conference on Computer Vision. pp. 3942--3951 (2021)

\bibitem{chexpert}
Irvin, J., Rajpurkar, P., Ko, M., Yu, Y., Ciurea-Ilcus, S., Chute, C., Marklund, H., Haghgoo, B., Ball, R., Shpanskaya, K., et~al.: Chexpert: A large chest radiograph dataset with uncertainty labels and expert comparison. In: Proceedings of the AAAI Conference on Artificial Intelligence. vol.~33, pp. 590--597 (2019)

\bibitem{johnson2019mimic}
Johnson, A.E., Pollard, T.J., Berkowitz, S.J., Greenbaum, N.R., Lungren, M.P., Deng, C.y., Mark, R.G., Horng, S.: Mimic-cxr, a de-identified publicly available database of chest radiographs with free-text reports. Scientific Data  \textbf{6}(1), ~317 (2019)

\bibitem{maskvlm}
Kwon, G., Cai, Z., Ravichandran, A., Bas, E., Bhotika, R., Soatto, S.: Masked vision and language modeling for multi-modal representation learning. In: The Eleventh International Conference on Learning Representations (2022)

\bibitem{mlip}
Li, Z., Yang, L.T., Ren, B., Nie, X., Gao, Z., Tan, C., Li, S.Z.: Mlip: Enhancing medical visual representation with divergence encoder and knowledge-guided contrastive learning. In: Proceedings of the IEEE/CVF Conference on Computer Vision and Pattern Recognition. pp. 11704--11714 (2024)

\bibitem{imitate}
Liu, C., Cheng, S., Shi, M., Shah, A., Bai, W., Arcucci, R.: Imitate: Clinical prior guided hierarchical vision-language pre-training. IEEE Transactions on Medical Imaging  (2024)

\bibitem{pneumo}
for imaging informatics~in medicine, S.: Siim-acr pneumothorax segmentation (2019), \url{https://www.kaggle.com/ c/siim-acr-pneumothorax-segmentation}

\bibitem{lovt}
M{\"u}ller, P., Kaissis, G., Zou, C., Rueckert, D.: Joint learning of localized representations from medical images and reports. In: European Conference on Computer Vision. pp. 685--701. Springer (2022)

\bibitem{oord2018representation}
Oord, A.v.d., Li, Y., Vinyals, O.: Representation learning with contrastive predictive coding. arXiv preprint arXiv:1807.03748  (2018)

\bibitem{covid}
Pavlova, M., Terhljan, N., Chung, A.G., Zhao, A., Surana, S., Aboutalebi, H., Gunraj, H., Sabri, A., Alaref, A., Wong, A.: Covid-net cxr-2: An enhanced deep convolutional neural network design for detection of covid-19 cases from chest x-ray images. Frontiers in Medicine  \textbf{9},  861680 (2022)

\bibitem{radford2021learning}
Radford, A., Kim, J.W., Hallacy, C., Ramesh, A., Goh, G., Agarwal, S., Sastry, G., Askell, A., Mishkin, P., Clark, J., et~al.: Learning transferable visual models from natural language supervision. In: International Conference on Machine Learning. pp. 8748--8763. PMLR (2021)

\bibitem{shih2019augmenting}
Shih, G., Wu, C.C., Halabi, S.S., Kohli, M.D., Prevedello, L.M., Cook, T.S., Sharma, A., Amorosa, J.K., Arteaga, V., Galperin-Aizenberg, M., et~al.: Augmenting the national institutes of health chest radiograph dataset with expert annotations of possible pneumonia. Radiology: Artificial Intelligence  \textbf{1}(1),  e180041 (2019)

\bibitem{nih}
Wang, X., Peng, Y., Lu, L., Lu, Z., Bagheri, M., Summers, R.M.: Chestx-ray8: Hospital-scale chest x-ray database and benchmarks on weakly-supervised classification and localization of common thorax diseases. In: Proceedings of the IEEE Conference on Computer Vision and Pattern Recognition. pp. 2097--2106 (2017)

\bibitem{wen2021rethinking}
Wen, Y., Chen, L., Deng, Y., Zhou, C.: Rethinking pre-training on medical imaging. Journal of Visual Communication and Image Representation  \textbf{78},  103145 (2021)

\bibitem{medklip}
Wu, C., Zhang, X., Zhang, Y., Wang, Y., Xie, W.: Medklip: Medical knowledge enhanced language-image pre-training. Proceedings of the IEEE/CVF International Conference on Computer Vision  (2023)

\bibitem{yang2021uncertainty}
Yang, S., Fevens, T.: Uncertainty quantification and estimation in medical image classification. In: International conference on artificial neural networks. pp. 671--683. Springer (2021)

\bibitem{convirt}
Zhang, Y., Jiang, H., Miura, Y., Manning, C.D., Langlotz, C.P.: Contrastive learning of medical visual representations from paired images and text. In: Machine Learning for Healthcare Conference. pp. 2--25. PMLR (2022)

\end{thebibliography}
\end{document}